\documentclass[lettersize,journal]{IEEEtran}
\usepackage{amsmath,amsfonts}
\usepackage{algorithmic}
\usepackage{algorithm}
\usepackage{array}
\usepackage[caption=false,font=normalsize,labelfont=sf,textfont=sf]{subfig}
\usepackage{textcomp}
\usepackage{stfloats}
\usepackage{url}
\usepackage{verbatim}
\usepackage{graphicx}
\usepackage{cite}

\usepackage{multirow}
\usepackage[normalem]{ulem}
\useunder{\uline}{\ul}{}
\begin{document}

\title{Domain generalization Person Re-identification on\\ Attention-aware multi-operation strategery}

\author{Yingchun Guo, Huan He, Ye Zhu, Yang Yu %,~\IEEEmembership{Staff,~IEEE,}
        % <-this % stops a space
\thanks{Yingchun Guo, Huan He, Ye Zhu, Yang Yu are with the School of Artificial Intelligence, Hebei University of Technology, Tianjin 300401, China(e-mail: {gyc@scse.hebut.edu.cn}, {202032803096@stu.hebut.edu.cn}, \{zhuye, yuyang\}@scse.hebut.edu.cn.}
\thanks{This work was supported in part by National Natural Science Foundation of China(61806071,62102129) and Natural Science Foundation of Hebei Province(F2019202381,F2019202464)  \textit{(corresponding author: Yang Yu.)}}
% <-this % stops a space
%\thanks{Manuscript received April 19, 2021; revised August 16, 2021.}
}

% The paper headers
\markboth{Journal of \LaTeX\ Class Files}%
{Guo \MakeLowercase{\textit{et al.}}: A Sample Article Using IEEEtran.cls for IEEE Journals}

% Remember, if you use this you must call \IEEEpubidadjcol in the second
% column for its text to clear the IEEEpubid mark.

\maketitle

\begin{abstract}
Domain generalization person re-identification (DG Re-ID) aims to directly deploy a model trained on the source domain to the unseen target domain with good generalization, which is a challenging problem and has practical value in a real-world deployment. In the existing DG Re-ID methods, invariant operations are effective in extracting domain generalization features, and Instance Normalization (IN) or Batch Normalization (BN) is used to alleviate the bias to unseen domains.  Due to domain-specific information being used to capture discriminability of the individual source domain, the generalized ability for unseen domains is unsatisfactory. To address this problem, an Attention-aware Multi-operation Strategery (AMS) for DG Re-ID is proposed to extract more generalized features. We investigate invariant operations and construct a multi-operation module based on IN and group whitening (GW) to extract domain-invariant feature representations. Furthermore, we analyze different domain-invariant characteristics, and apply spatial attention to the IN operation and channel attention to the GW operation to enhance the domain-invariant features. The proposed AMS module can be used as a plug-and-play module to incorporate into existing network architectures. Extensive experimental results show that AMS can effectively enhance the model's generalization ability to unseen domains and significantly improves the recognition performance in DG Re-ID on three protocols with ten datasets.
\end{abstract}


\section{Introduction}
\IEEEPARstart{T}{he} person Re-Identification (Re-ID) task refers to detecting and matching whether persons are the same one captured by different cameras in a non-overlapping camera network. In recent years, due to the rapid development of security and intelligent monitoring systems, Re-ID has received more and more attention. The main challenge of Re-ID is the variation in different cameras, such as body pose, viewing angle, lighting, image resolution, occlusion, background, and so on. Generally, Re-ID can be viewed as a subproblem of image retrieval to quickly and accurately match the query image from a large-scale gallery.

To address these challenges, researchers propose a large amount of fully supervised person Re-ID methods\cite{zhou2017large,wei2018glad,dai2019batch,chen2020salience}. These methods achieve satisfactory performance when training and testing are performed on the same dataset. However, when transferred to a new domain, due to the domain bias, the performance degrades drastically on different domains. This reveals a lack of generalization ability in single-dataset supervised models since the learned features overfit the training domain instead of capturing general features relevant to person discrimination. Therefore, some researchers have focused on unsupervised domain adaptation (UDA), which is more practical than supervised methods. The UDA methods use the unlabeled data collected from the target domain to alleviate the domain overfitting problem\cite{huang2018eanet, chen2019instance,huang2019domain, li2021intra}. However, the target domain data is still required, and sometimes the data collection of the target domain is expensive or even impossible. Recently, researchers turned to Domain Generalization (DG) methods, which utilize the labeled data from multiple source domains to learn new generalizable models for unseen target domains without using any target domain data for training\cite{segu2020batch, mansilla2021domain, kim2021selfreg, pandey2021generalization}. DG is more practical than the supervised methods and UDA. In this paper, we focus on the DG Re-ID problem to extract domain invariant features and learn a model that performs well for unseen domains.

DG Re-ID methods adopt invariant operations\cite{jin2020style, liu2022debiased}, meta-learning\cite{dai2021generalizable,bai2021person30k}, data augmentation\cite{zhou2021domain, zhao2021learning} or other methods\cite{luo2020generalizing, yang2022domain,chen2021dual} to obtain generalizable domain-invariant features. Among them, the invariant operation is proved to be the most effective method\cite{pan2019switchable, choi2021meta}. Through invariant operations, such as Batch Normalization (BN)\cite{ioffe2015batch}, Instance Normalization (IN)\cite{ulyanov2016instance}, or their combination, DG Re-ID methods attempt to get domain generalization features. In these domain-invariant operations, IN is used to capture and eliminate style variations among different domains\cite{pandey2021generalization}, and BN can improve discriminative ability on features by standardizing input data or activations\cite{liu2022debiased}, so the combination of IN and BN is generally used to improve the generalization performance\cite{nam2018batch} . However, it is hard work to balance BN and IN, which makes the domain-invariant features are still not generalizable. Furthermore, domain-specific information is still used in DG Re-ID, which makes the source domain to be fitted better, while weakening the generalization ability to unknown domains. A whitening operation is a technique that removes feature correlation and makes each feature have unit variance. It has been proven that the feature whitening can effectively eliminate domain-specific style information in image translation\cite{cho2019image}, style transfer\cite{li2017universal}, and domain adaptation\cite{pan2019switchable, sun2016deep, roy2019unsupervised}, and thus it may improve the generalization ability of the feature representation\cite{choi2021robustnet}, but not yet fully explored in DG Re-ID.

We focus on the invariant operations by investigating their role in extracting general domain features and using them to address overfitting to the source domain. Since IN can well remove instance-level style information, such as illumination, contrast, and so on\cite{ulyanov2016instance}, it is used to remove style information in this paper. Different from the early invariant operation combination, we use group whitening (GW)\cite{huang2021group} to replace BN as the invariant operation. Although the stochasticity of normalization introduced along the batch dimension in BN operation is believed to benefit generalization, it also results in differences between the training distribution and the test distribution, meanwhile, BN's error increases rapidly as the batch size becomes smaller. Since GW operation divides the neurons of a sample into groups for standardization over the neurons in each group and then decorrelates the groups, which has stable performance for a wide range of batch sizes. Therefore, based on IN and GW operations, we propose an attention-aware multi-operation strategery(AMS) to extract domain-invariant feature representations. In AMS, we concatenate GW with IN in an ingenious combination and then apply spatial attention and channel attention to these two operation modules respectively. Through extensive experiments on IN and GW operations, we verify that IN can process the spatial information at the instance level, while the GW operation can process information in the channel dimension. Furthermore, ASM with spatial attention and channel attention is inserted into the existing network architectures to extract more domain-invariant features. Experiments in unseen domains demonstrate the effectiveness of the proposed method.
Our contributions are as follows:

1) We propose a simple but effective plug-and-play attention-aware multi-operation strategery to achieve domain-invariant features.

2) We construct a multi-operation module based on IN and GW to extract domain-invariant feature representations, and further apply spatial attention and channel attention to IN and GW to enhance domain-invariant features.

3) We evaluate the proposed method on eight standard benchmark datasets, and the experimental results show AMS outperforms most of the state-of-the-art methods. 

The remainder of this paper is organized as follows: the related work is presented in Section 2; the proposed method is presented in Section 3; the experimental results and analysis are arranged in Section 4, and the conclusion is given in Section 5.

\section{Related Word}
\subsection{Domain Generalization Person Re-Identification}
According to the number of source domain datasets, existing methods can be classified into source domain with a single dataset\cite{zhou2021learning, jin2020style,liao2020interpretable} and source domain with multiple large-scale datasets\cite{song2019generalizable,chen2021dual,luo2020generalizing}. Since multiple source domains can better reflect the generalization ability of the model to unknown domains, the existing DG Re-ID mainly focuses on the multi-source DG problem. For DG Re-ID based on multi-source datasets, one effective solution is meta-learning. Song \textit{et al}.\cite{song2019generalizable}  conducted the first attempt at applying meta-learning to DG Re-ID. They divided the training set into gallery and probe and simulated the test scene matching the probe image with the gallery set, resulting in a generalizable model. Dai \textit{et al}.\cite{dai2021generalizable} divided the training set into meta-training sets and meta-test sets to simulate the generalization process of an unknown domain, learned a voting network that can score expert networks in different domains, and finally, a linear combination of experts in the source domain is used to simulate the characteristics of the unknown domain. In addition to meta-learning, data augmentation is also used to solve DG Re-ID, which mainly includes mixup and generation using networks. Luo \textit{et al}.\cite{luo2020generalizing} mixuped images from different domains as an intermediate domain, which avoided the drastic bias caused by abrupt transitions between two very different domains. Zhou \textit{et al}.\cite{zhou2020learning} used a data generator trained with optimal transport to synthesize data from pseudo-new domains to enhance the diversity of available training domains. Considering that the meta-training process of the meta-learning method is complex and slow, which limits its application in DG Re-ID; at the same time, the data augmentation method is only expansion of the existing category data, and cannot generate unknown data, which would lead to overfitting to the known domains. Different from these methods, there is a way to directly extract domain-invariant features, namely, invariance techniques, which have recently received more and more attention. 

\subsection{Invariant methods}
Since the key to DG Re-ID is to obtain domain-invariant features that generalize well to unknown domains, invariance technology has attracted more and more attention. The invariance technology can optimize and extract domain-invariant features and has been used in different deep neural network architectures. The BN operation\cite{ioffe2015batch} is most widely used in invariance technology, such as domain-specific BN\cite{liu2022debiased}, Batch-Instance Normalization  (BIN)\cite{choi2021meta}, and so on\cite{li2017universal,huang2021group}. The BN operation\cite{ioffe2015batch}  normalizes the inputs by using statistics computed in mini-batches, and regularizes feature representations from heterogeneous domains, therefore the trained model is often capable of adapting to unseen domains. Liu \textit{et al}.\cite{liu2022debiased}  constructed a Gaussian process-based framework for domain-specific BN to estimate potential new domains and effectively improved the generalization ability of the model on an unseen domain. However, domain-specific BNs contain the inherent flaw that includes domain-specific information, when domain shift is significant, we often need to remove domain-specific styles for better generalization. The IN operation\cite{ulyanov2016instance}, which performs BN-like computations on a single sample, can remove domain-specific style information\cite{zhou2021domain}  in the sample. Jia \textit{et al}.\cite{jia2019frustratingly}  built a baseline of DG Re-ID by inserting IN\cite{ulyanov2016instance} in each bottleneck in the shallow layers of the deep network, and obtained robust domain-invariant features over multiple unknown domains. The IN operation is an effective scheme incorporating both batch and instance normalization techniques to further improves accuracy, however, its operation on the standard deviation cannot reflect the image style information well. Recent studies have found that the covariance encodes the image style information better than the standard deviation\cite{li2017universal}, therefore some researchers turn to process covariance information, that is, whitening operations, such as Instance Whitening(IW)\cite{li2017universal},  Group Whitening(GW) \cite{huang2021group}, Switchable Whitening(SW)\cite{pan2019switchable}.

Inspired by these methods, we propose a multi-operation module based on IN and GW operations to extract efficient domain-invariant features of DG Re-ID.

\subsection{Difference from previous works}
The normalization techniques we explore have been abundantly presented in existing works. The contributions of our work lie in the following aspects: (1) we explored the feasibility of applying multiple invariant operations to the inverted residual bottleneck in ResNet50, and verified that the combination of IN followed by GW can achieve the optimal effect to overcome Re-ID in domain-specific information; (2) we systematically demonstrate that both normalization techniques are biased towards spatial and channel information and AMS provides the strongest generalization ability for Person Re-ID while being simple and requiring no target domain data.

\section{Method}
In this work, our goal is to learn domain-invariant feature representations that can be generalized well. The overall framework of the proposed method is shown in Fig.1. We use $K$-labeled source domains as training sets, denoted as $D=\{D\}_{k=1}^K$, where $D_k=\{x_n^k,y_n^k\}_{n=1}^{N_k}$, $N_k$ represents the total number of images in the $k$-th source domain. Each sample $x_n^k$ has a label as $y_n^k\in\{1,2,\cdots,M_k\}$, where $M_k$ represents the total identity number of the source domain $D_k$. The label set of each source domain is different from others, and the number of identities of all source domains can be expressed as $M=\sum_{k=1}^K {M_k}$.

\begin{figure*}[!t]
\centering
\includegraphics[width=.7\textwidth]{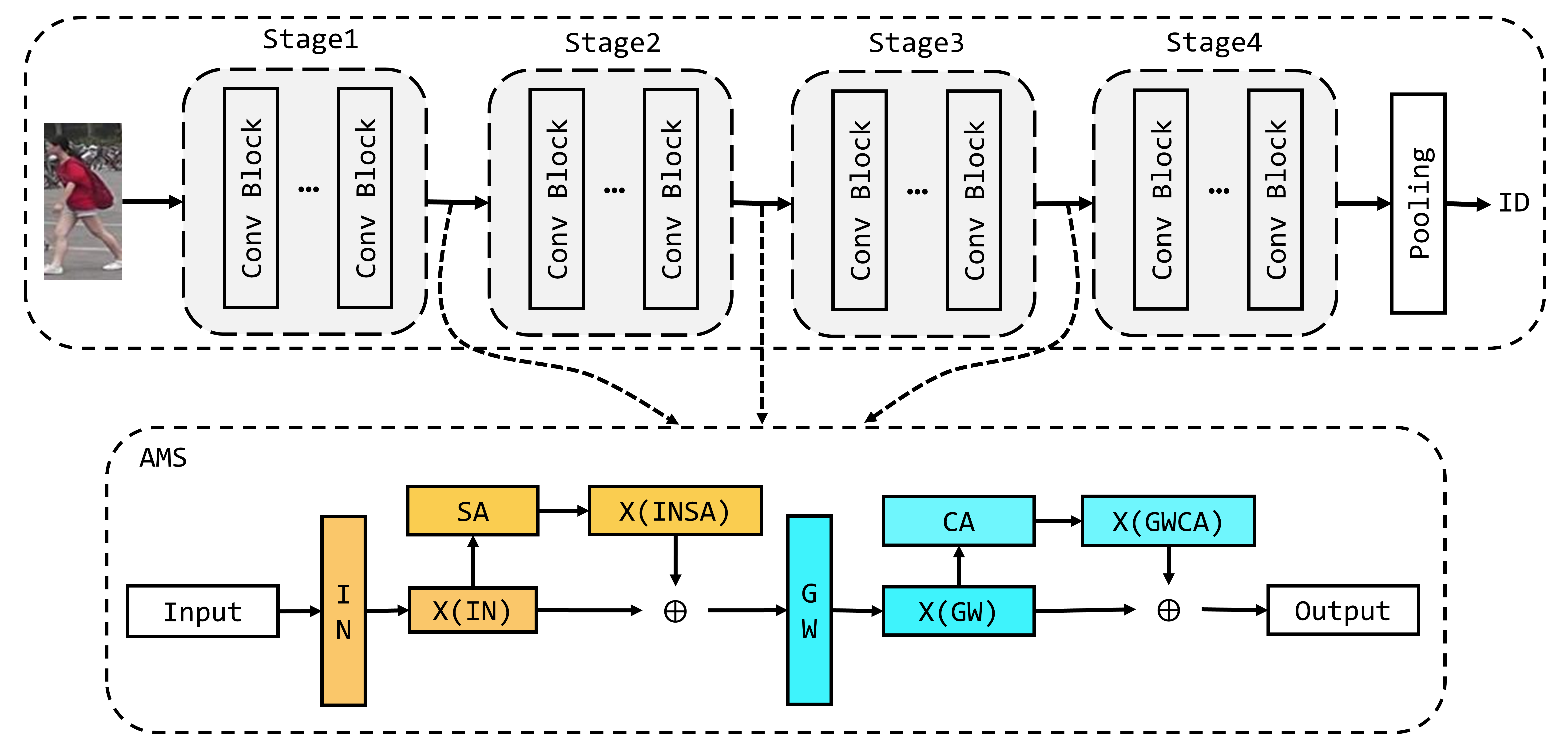}
\caption{The framework of this paper and AMS is inserted into the first three bottleneck layer of ResNet50. In this paper, the backbone network is ResNet50, and ASM is inserted into the first three layers. For each AMS module, the input is the output of the bottleneck layer. The AMS uses the output of the bottleneck layer as the module input. For the IN operation, the instance normalized feature X(IN) adds its spatial attention (SA) feature X(INSA) to feed into the GW operation. Furthermore, for the GW operation, the group-whitened feature X(GW) adds its channel attention (CA) feature X(GWCA) to output to the main network. $\bigoplus$ means element-wise addition.}
\label{fig_1}
\end{figure*}

During the training phase, we aggregate images from all source domains to train the DG model. AMS is inserted in the first three bottleneck layers of the backbone network (i.e., ResNet50).   Following our DG goal, we employ the IN techniques to generalize well to unseen domains. Unlike other schemes, we design the AMS module, which uses IN to remove the instance-level style information, then uses GW to remove the channel-wise style information. Meanwhile, to enable this double regularization technique to extract more generalization features, we apply instance-level spatial attention to IN and channel-wise channel attention to GW.

\subsection{Instance Normalization}
Since IN can effectively remove the illumination and contrast information in the image\cite{pan2019switchable}, it is used in the proposed module to extract domain-invariant features. Given a mini-batch feature map $f \in \mathbb{R}^{B\times C\times H\times W}$, where \textit{B}, \textit{C}, \textit{H} and \textit{W} are the number of samples, channels, height, and width respectively. Here $b\in B$, $c\in C$, $h\in H$, and $w\in W$ denote the indices of each dimension respectively. We normalize the feature using feature statistics (i.e. mean and standard deviation) for each channel of each sample, and we can get the instance-normalized feature $f$ with Eq.(\ref{E1}) as follows:

\begin{equation}
\label{E1}
    {\rm IN}(f)=\gamma \frac{f-\mu(f)}{\sigma(f)+\epsilon} + \beta 
\end{equation}
where $\gamma, \beta \in \mathbb{R}^C$ is a learnable affine transformation parameter, and $\epsilon$ is a small bias to prevent the denominator from tending to 0. $\mu(f), \sigma(f) \in \mathbb{R}^{B \times C}$ are the mean and standard deviation, corresponding to the spatial dimension of each channel for each instance $\mu (f)_{b,c}$ and $ \sigma(f)_{b,c}$, as shown in Eq.(\ref{E2}) and Eq.(\ref{E3}):

\begin{equation}
\label{E2}
    \mu(f)_{b,c}=\frac{1}{HW}\sum_{h=1}^{H}\sum_{w=1}^{W}f_{b,c,h,w}
\end{equation}
where  $f_{b,c,h,w}$ is the feature value, which belongs to $c$-th channels, $h$-th row, and $w$-th column of $b$-th sample.

\begin{equation}
\label{E3}
    \sigma(f)_{b,c}=\sqrt{\frac{1}{HW}\sum_{h=1}^{H}\sum_{w=1}^{W}(f_{b,c,h,w}-\mu(f)_{b,c})^2}
\end{equation}
where $b\in B$, $c \in C$, $h \in H$, $w \in W$ is the index of each dimension, respectively.

Recent studies\cite{jia2019frustratingly,pan2019switchable}  have shown that style information is stored in the bottom layer of convolutional neural networks through instance-level feature statistics, while classification information\cite{zhou2021domain} is stored in high layers. Hence, we adopt the same setting as Zhou \textit{et al}. \cite{zhou2021domain}, that is inserting the AMS module with IN into the first three stages of ResNet-50\cite{he2016deep} to extract domain-invariant features.

\subsection{Group Whitening}
From Pan's research\cite{pan2019switchable}, we know that whitening operation can filter style information, such as appearance and color. Therefore, to further extract domain-invariant features, we apply a whitening operation following the IN operation, extending the first-order standard deviation information of the normalization operation to the second-order covariance information of the whitening operation. As for a kind of whitening operation, Instance Whitening(IW)\cite{li2017universal} can unify the joint distribution and remove the correlation between channels, which is an effective operation to extract domain-invariant features. However,  IW is a strong whitening operation that would strictly remove the correlation among all channels and corrupt the semantic content, furthermore resulting in the loss of invariant information in the key domains. To alleviate this problem, we remove channel correlation in each group by grouping the channels, i.e. we use GW instead of IW. The specific implementation of GW is as follows.

Given a mini-batch feature map $f\in{\mathbb{R}}^{B\times C \times H \times W}$, where \textit{B}, \textit{C}, \textit{H} and \textit{W} is the number of samples, number of channels, height, and width. The small batch feature map is divided into $g$ groups according to the channel dimension, and each group has $c$ channels, where $C=g\times c$. We transform the feature map shape ${\mathbb{R}}^{B\times C \times H \times W}$ into shape ${\mathbb{R}}^{B\times g \times cHW}$ as $f_G$ is:

\begin{equation}
\label{E4}
    f_G=\prod(f;g)\in{\mathbb{R}}^{B\times g \times cHW}
\end{equation}
where $\prod(\cdot;g)$ is the group division operation. The corresponding mean and covariance are calculated as:

\begin{equation}
\label{E5}
    \mu_G=\frac{1}{cHW}f_G\times {\bf{1}}
\end{equation}
\begin{equation}
\label{E6}
    {\sum}_{G}=\frac{1}{cHW}(f_G-\mu_G\times 1^T)(f_G-\mu_G\times 1^T)^T+\varepsilon {\bf{I}}
\end{equation}
where {\bf{1}} is a column vector of all ones, and $\varepsilon$ represents a small positive number to prevent numerical instability. {\bf{I}} is the identity matrix. And the obtained whitening feature is:

\begin{equation}
\label{E7}
    \hat{f}_G={{\sum}_{G}}^{-\frac{1}{2}}(f_G-\mu_G\times 1^T) \in {\mathbb{R}}^{B\times g \times cHW}
\end{equation}

The final output feature is:

\begin{equation}
\label{E8}
   	F={\prod}^{-1}(\hat{f}_G)\in{\mathbb{R}}^{B\times C \times H \times W } 
\end{equation}
	
In this paper, we set the hyperparameter \textit{g} to 64, and the detailed comparison experiments are in Section 4.5.

\subsection{Attention-aware module}
Based on different domain-invariant characteristics, IN is used to normalize the spatial dimension of each channel for each sample and seldom to normalize the feature channel-wise\cite{ulyanov2016instance}; while GW is applied to divide channels into several groups and normalize these groups to focus on the correlation among channels\cite{peng2022semantic}. Since IN has normalized the spatial information before GW, the GW operation can get more specific information to the channel information. Therefore, it can be inferred that IN focuses on domain-invariant spatial features, while GW focuses on domain-invariant channel features. Furthermore, spatial attention and channel attention are applied to instance-normalized features and group-whitened features respectively to enhance these features. The comparison experiments with different spatial and channel attention applied to IN and GW are given in Table 4. The specific implementation of the attention module is given as follows.

Suppose a mini-batch feature map $f\in{\mathbb{R}}^{B\times C \times H \times W}$, where \textit{B}, \textit{C}, \textit{H} and \textit{W} are the number of samples, the number of channels, the height, and width, and the output of the IN module is:

\begin{equation}
\label{E9}
    f^{'}_{\rm IN}=f_{\rm IN}+f_{\rm SA}
\end{equation}

\begin{equation}
\label{E10}
    f_{\rm IN}={\rm IN}(f)
\end{equation}

\begin{equation}
\label{E11}
    f_{\rm SA}={\rm SA}(f_{\rm IN})
\end{equation}
where ${\rm IN}(\cdot)$ represents the IN operation, and ${\rm SA}(\cdot)$ represents the spatial attention module(Convolutional Block Attention Module(CBAM)\cite{woo2018cbam} is used in this paper).

Then, we take the output of the IN module as the input of the GW module, and the output of the GW module is:

\begin{equation}
\label{E12}
    f^{'}=f_{\rm GW}+f_{\rm CA}
\end{equation}

\begin{equation}
\label{E13}
    f_{\rm GW}={\rm GW}(f^{'}_{\rm IN})
\end{equation}

\begin{equation}
\label{E14}
    f_{\rm CA}={\rm CA}(f_{\rm GW})
\end{equation}
where, ${\rm GW}(\cdot)$ represents the GW operation, and ${\rm CA}(\cdot)$ represents the channel attention module(CBAM\cite{woo2018cbam} is used in this paper).

Take $f^{'}$ as the output of the AMS module and feed it back to the network to continue forward propagation. The flowchart of the algorithm is shown in Algorithm 1.

\begin{algorithm}[H]
\renewcommand{\algorithmicrequire}{\textbf{Input:}}
\renewcommand{\algorithmicensure}{\textbf{Output:}}
\caption{AMS.}\label{alg:alg1}
\begin{algorithmic}[1]
    \REQUIRE $f\in{\mathbb{R}}^{C \times H \times W}$  through the bottleneck layer
	\ENSURE $f^{'}\in{\mathbb{R}}^{C \times H \times W}$ after the AMS module
    \STATE Compute IN feature: $f_{\rm IN}$ 
    \STATE Apply Spatial Attention to the IN feature: $f_{\rm SA}={\rm SA}(f_{\rm IN})$
    \STATE $f^{'}_{\rm IN}=f_{\rm IN}+f_{\rm SA}$
    \STATE Compute GW feature: $f_{\rm GW}={\rm GW}(f^{'}_{\rm IN})$ 
    \STATE Apply Channel Attention to the GW feature: $f_{\rm CA}={\rm CA}(f_{\rm GW})$
    \STATE $f^{'}=f_{\rm GW}+f_{\rm CA}$

\end{algorithmic}
\label{alg1}
\end{algorithm}

\begin{figure*}[!t]
\centering
\includegraphics[width=.7\textwidth]{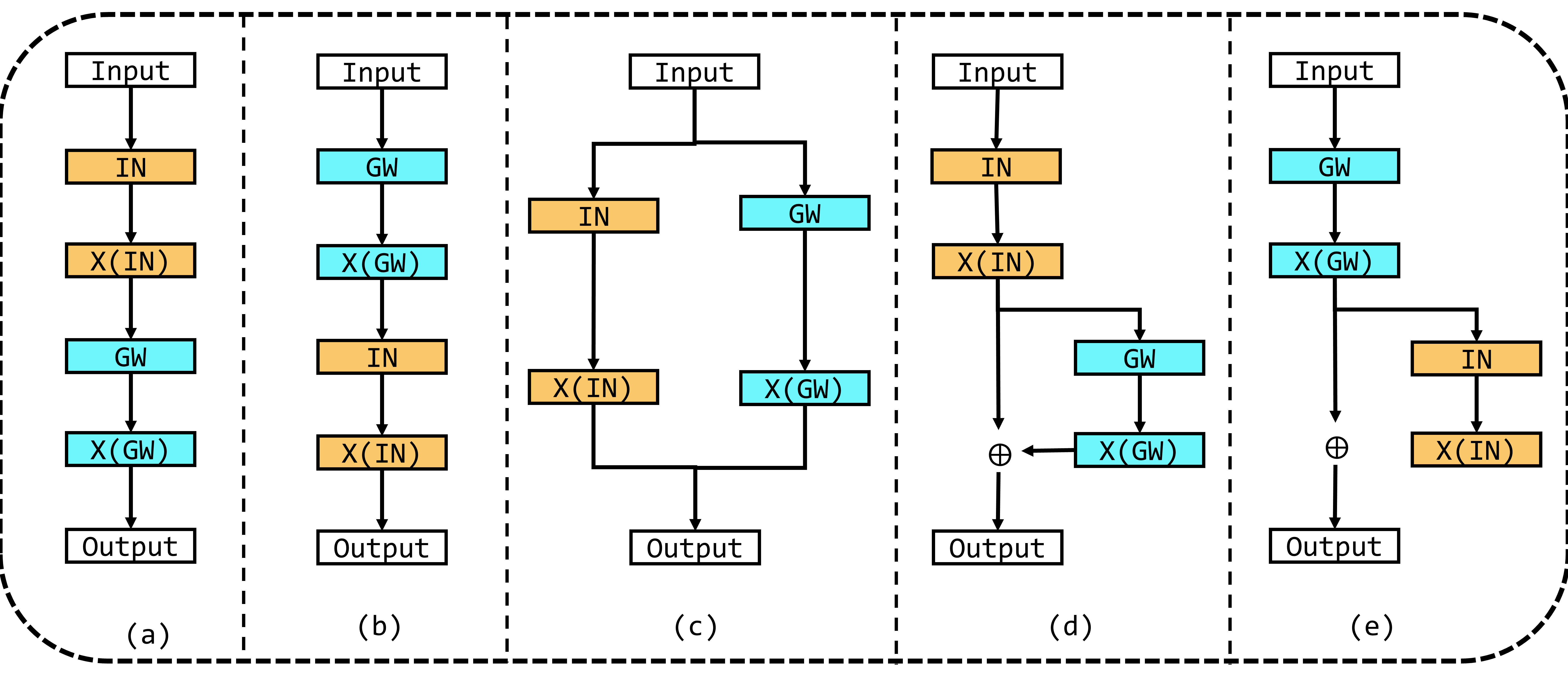}
\caption{Schematic diagram of different multi-operation modules. (a) represents the tandem of IN and GW operations; (b) represents the tandem of GW and IN operations; (c) represents the parallel connection of IN and GW operations; (d) represents the instance normalized feature is added together with its feature further applied with GW operation; (e) is that the feature applied with GW is added together with its feature further applied with IN operation.}
\label{fig_2}
\end{figure*}

\subsection{Loss Function}
During training, we use cross-entropy loss ${\mathcal{L}_{cls}}$ and triplet loss ${\mathcal{L}_{tri}}$ for model training. The cross-entropy loss is defined as:
	
\begin{equation}
\label{E15}
    {\mathcal{L}_{cls}}=-\sum_{i=1}^{N}log\frac{exp(p)}{\sum_{c=1}^{C}exp({p}_i)}
\end{equation}
where $N$ and $C$ represent the number of samples and the channel dimension of the classification feature, respectively, and $p\in \mathbb{R}^{N\times C}$ is the classification feature.
	
The triplet loss is defined as:
\begin{equation}
\label{E16}
    {\mathcal{L}_{tri}}=\sum_{i=1}^{N} {\left[  d_i^p  -  d_i^n + \alpha \right]}_+
\end{equation}
where $N$ denotes the number of samples, $d^p$ and $d^n$ represent the Euclidean distance between the anchor sample and its hardest positive and hardest negative samples. $\left[\cdot\right]_+$ represents the hinge function $max(\cdot,0)$, $\alpha$ represents the margin, which is set to 0.3, following most methods.
	
The total loss for model training can be defined as:

\begin{equation}
\label{E17}
    {\mathcal{L}_{overall}}={\mathcal{L}_{cls}} + \lambda_{tri} {\mathcal{L}_{tri}}
\end{equation}
where $\lambda_{tri}$ is the balance weights.

\subsection{Different multi-operation strategy comparisons}
In addition to the multi-operation module that IN layer is followed by GW, we also perform extensive operation combinations, as shown in Fig.2. In module (b), the GW operation is performed first, and then is the IN operation. In Module (c), the IN and GW operations are performed at the same time and then added together to output. In Module (d), the invariant operation IN is taken as the first operation, and then GW is the second operation, and the output of this module equals sum of Module (a) and the IN operation. The output of Module(e) equals sum of Module (b) and the GW operation. A detailed comparative experiment is given in Section 4.4.

%\begin{figure*}[!t]
%\centering
%\includegraphics[width=.7\textwidth]{module.png}
%\caption{Schematic diagram of different multi-operation modules. (a) represents the tandem of IN and GW operations; (b) represents the tandem of GW and IN operations; (c) represents the parallel connection of IN and GW operations; (d) represents the instance normalized feature is added together with its feature further applied with GW operation; (e) is that the feature applied with GW is added together with its feature further applied with IN operation.}
%\label{fig_2}
%\end{figure*}

\section{Experiments}
\subsection{Dataset and evaluation settings}
We evaluate the effectiveness of the proposed AMS on multiple protocols. For DG Re-ID, protocol refers to how source and target domain datasets are organized for model performance evaluation. Different evaluation settings are shown in Table 1.

\textbf{Protocol-1:} Following the existing methods, we employ common-used person Re-ID benchmarks to evaluate the generalization ability, where five large-scale datasets are used as the source domain, and four small-scale datasets are used as invisible target domains. As shown in Table 2, the source domains include CUHK02\cite{li2013locally}, CUHK03\cite{li2014deepreid}, Market-1501\cite{zheng2015scalable}, DukeMTMC-reID\cite{zheng2017unlabeled} and CUHK-SYSU\cite{xiao2017joint}. Target domains include VIPeR\cite{gray2008viewpoint}, PRID\cite{hirzer2011person}, GRID\cite{loy2009multi} and iLIDS\cite{zheng2009associating}. All training sets and test sets in the source domain are used to train the model. In the testing phase, the performance of the model on four small-scale per-son Re-ID datasets was tested separately, and the final performance was averaged by performing 10 repeated random splits on the test set.

\textbf{Protocol-2 and Protocol-3:} Considering the poor image quality of small-scale person Re-ID datasets, the performance on these datasets cannot accurately reflect the generalization ability of the model in real-world scenarios. Zhao \textit{et al}.\cite{zhao2021learning} and Dai \textit{et al}.\cite{dai2021generalizable} set a new protocol (i.e. leave-one-out setting) for four large-scale person Re-ID datasets. Specifically, four large person Re-ID datasets (Market-1501\cite{zheng2015scalable}, DukeMTMC-reID\cite{zheng2017unlabeled}, CUHK03\cite{li2014deepreid}, MSMT17\cite{wei2018person}) are divided into two parts: three datasets are used as the source domain for training, and the remaining one is used as the target domain for testing. The scheme is divided into Protocol-2\cite{zhao2021learning} and Protocol-3\cite{dai2021generalizable} according to whether the source domain uses both the training set and the test set. For simplicity, in the following sections, we denote Market1501 as M, DukeMTMC-reID as D, CUHK02 as C2, CUHK03 as C3, MSMT17 as MT, and CUHK-SYSU as CS.

\textbf{Evaluation Setting:} We evaluate the performance of person Re-ID methods using mean Average Precision (mAP) and Cumulative Matching Characteristic (CMC) of Rank-k, which are common-used in DG Re-ID.

\begin{table}[h]
\tabcolsep=6pt
\centering
\caption{Different evaluation Protocols.}
\small
    \begin{tabular}{c|ll}
    	\hline
    	Setting   & \multicolumn{1}{c|}{Training Set}          & \multicolumn{1}{c}{Testing Set}  \\ \hline
    	Protocol-1 & \multicolumn{1}{l|}{M+D+C2+C3+CS} & PRID, GRID, VIPeR, iLIDs \\ \hline
    	Protocol-2 & \multicolumn{2}{l}{Leave-one-out for M+D+C3+MT, train}                     \\ \hline
    	Protocol-3 & \multicolumn{2}{l}{Leave-one-out for M+D+C3+MT, combat}                    \\ \hline
    \end{tabular}
  \label{tab:dataset}
\end{table}
\subsection{implementation details}
We use pretrained ResNet-50\cite{he2016deep} on ImageNet\cite{deng2009imagenet} as the backbone network, following the state-of-the-art(SOTA) methods\cite{luo2019strong, zhao2021learning, liu2022debiased}, the last residual layer's stride size is set to 1. Following the backbone, we use a generalized mean pooling layer\cite{radenovic2018fine} with a batch normalization layer as a baseline to extract person Re-ID features. All input images size is resized to $384\times 128$, and the batchsize is set to 128, including 8 identities, each with 16 images. For data augmentation, we employ random horizontal flipping, random cropping, color dithering, and random erasing. We set an initial learning rate of $3.5\times 10^{-4}$, use a warmup policy for the first 10 epochs, and decay to $7.7\times 10^{-7}$  with cosine annealing strategy in subsequent epochs. The weights for cross-entropy loss and triplet loss are set equal, i.e, $\lambda_{tri}=1$. We use Adam as the optimizer and the weight decay is set to $5\times 10^{-4}$. The total training stage takes 60 epochs. To speed up the training process and improve efficiency, we use automatic mixed precision training\cite{micikevicius2017mixed}. All experiments are performed on an NVIDIA GeForce RTX 3090 GPU.

\subsection{Comparisons with the SOTA methods}
The proposed method is compared with a wide range of existing methods, including the following fields:

\textbf{General meta-learning methods:} predicting the parameters from the activations(PPA)\cite{qiao2018few} and first-order gradient-based meta-learning algorithm(Reptile)\cite{nichol2018first}.

\textbf{General domain generalization method:} meta-learning domain generalization(MLDG)\cite{li2018learning} and cross-gradient training(CrossGrad)\cite{shankar2018generalizing}. 

\textbf{Domain aggregation methods:} Aggregation Part-based Convolutional Base-line(Agg\_PCB)\cite{sun2018beyond} and Aggregation Aligned ReID(Agg\_Align)\cite{zhang2017alignedreid}. 

\textbf{Dedicated methods for person re-identification:} Bag of Tricks(BoT)\cite{luo2019strong}, Beginner Classifier as Regularization(BCaR)\cite{tamura2020bcar} and Dual Distribution Alignment Network(DDAN)\cite{chen2021dual}.

\textbf{Modules and Architecture Design Methods:} Style Normalization and Restitution(SNR)\cite{jin2020style} and Query-Adaptive Convolution, QAConv$_50$)\cite{liao2020interpretable}.

\textbf{Meta-Learning Methods:} Domain-Invariant Mapping Network(DIMN)\cite{song2019generalizable}, Dual-Meta Generalization Network(DMG-Net)\cite{bai2021person30k} and relevance-aware mixture of experts(RaMoE)\cite{dai2021generalizable}. 

\textbf{Invariance research methods:} Dual Normalization(DualNorm)\cite{jia2019frustratingly}, Meta Batch-Instance Normalization(MetaBIN)\cite{choi2021meta}, Memory-based Multi-Source Meta-Learning(M$^{3}$L)\cite{zhao2021learning} and Debiased Batch Normalization via Gaussian Process(GDNorm)\cite{liu2022debiased}.

The existing DG Re-ID methods do not use invariant methods or only use instance normalization without considering further whitening operations. Compared with them, AMS can significantly improve the generalization ability. To verify the effectiveness of AMS, we set up three comparative experiments, which are compared on Protocol-1, Protocol-2 and Protocol-3, as shown in Tables 2 and 3.

\textbf{Comparison under Protocol-1}. As shown in Table 2, we compare our method with the SOTA methods under Protocol-1, and our method achieves superior performance against other methods. Specifically, on the average performance of the four test sets, our method outperforms GDNorm\cite{liu2022debiased} by 0.41\% mAP and 0.32\% R1 accuracy. Since GDNorm uses domain-specific BN to model Gaussian processes in the unknown domains, ignoring the exploration of invariant methods. For MetaBIN\cite{choi2021meta}, our mAP and R1 accuracy rates are 3.03\% and 3.22\% higher, respectively, the reason is that MetaBIN only takes the balance between IN and BN for consideration, while our method is more capable of extracting domain-invariant features. Our method outperforms RaMoE by 7.58\% mAP and 7.64\% R1 accuracy, since RaMoE uses domain experts to simulate unknown domains, and the extracted features may not have good generalization.

\begin{table*}[ht]
\tabcolsep=4pt
\small
\centering
    \caption{Comparison of performance with the SOTA methods under Protocol-1(\%)}
\begin{tabular}{l|cc|cllc|cllc|cllc|cllc}
\hline
\multicolumn{1}{c|}{\multirow{2}{*}{Method}} & \multicolumn{2}{c|}{Average}  & \multicolumn{4}{c|}{VIPeR}                                                          & \multicolumn{4}{c|}{PRID}                                                           & \multicolumn{4}{c|}{GRID}                                                           & \multicolumn{4}{c}{i-LIDS}                                                                 \\ \cline{2-19} 
\multicolumn{1}{c|}{}                        & R1           & mAP           & R1           & \multicolumn{1}{c}{R5}  & \multicolumn{1}{c}{R10} & mAP           & R1           & \multicolumn{1}{c}{R5}  & \multicolumn{1}{c}{R10} & mAP           & R1           & \multicolumn{1}{c}{R5}  & \multicolumn{1}{c}{R10} & mAP           & R1           & \multicolumn{1}{c}{R5}  & \multicolumn{1}{c}{R10} & mAP                  \\ \hline
Agg\_Align\cite{zhang2017alignedreid}                                & 34.9          & 44.5          & 42.8          &                          &                          & 52.9          & 17.2          &                          &                          & 25.5          & 15.9          &                          &                          & 24.7          & 63.8          &                          &                          & 74.7                 \\
Reptile\cite{nichol2018first}                                  & 28.1          & 37.1          & 22.1          &                          &                          & 31.3          & 17.9          &                          &                          & 26.9          & 16.2          &                          &                          & 23.0          & 56.0          &                          &                          & 67.1                 \\
GrossGrad\cite{shankar2018generalizing}                                & 24.6          & 34.0          & 20.9          &                          &                          & 30.4          & 18.8          &                          &                          & 28.2          & 9.0           &                          &                          & 16.0          & 49.7          &                          &                          & 61.3                 \\
Agg\_PCB\cite{sun2018beyond}                                  & 40.6          & 49.0          & 38.1          &                          &                          & 45.4          & 21.5          &                          &                          & 32.0          & 36.0          &                          &                          & 44.7          & 66.7          &                          &                          & 73.9                 \\
MLDG\cite{li2018learning}                                    & 29.3          & 39.4          & 23.5          &                          &                          & 33.5          & 24.0          &                          &                          & 35.4          & 15.8          &                          &                          & 23.6          & 53.8          &                          &                          & 65.2                 \\
PPA\cite{qiao2018few}                                      & 42.1          & 52.6          & 45.1          &                          &                          & 54.5          & 31.9          &                          &                          & 45.3          & 26.9          &                          &                          & 38.0          & 64.5          &                          &                          & 72.7                 \\
DualNorm\cite{jia2019frustratingly}                                 & 57.6          & 61.8          & 53.9          & \multicolumn{1}{c}{62.5} & \multicolumn{1}{c}{75.3} & 58.0          & 60.4          & \multicolumn{1}{c}{73.6} & \multicolumn{1}{c}{84.8} & 64.9          & 41.4          & \multicolumn{1}{c}{47.4} & \multicolumn{1}{c}{64.7} & 45.7          & 74.8          & \multicolumn{1}{c}{82.0} & \multicolumn{1}{c}{91.5} & 78.5                 \\
DIMN\cite{song2019generalizable}                                     & 47.5          & 57.9          & 51.2          & \multicolumn{1}{c}{70.2} & \multicolumn{1}{c}{76.0} & 60.1          & 39.2          & \multicolumn{1}{c}{67.0} & \multicolumn{1}{c}{76.7} & 52.0          & 29.3          & \multicolumn{1}{c}{53.3} & \multicolumn{1}{c}{65.8} & 41.1          & 70.2          & \multicolumn{1}{c}{89.7} & \multicolumn{1}{c}{94.5} & 78.4                 \\
SNR\cite{jin2020style}                                      & 57.3          & 66.4          & 52.9          &                          &                          & 61.3          & 52.1          &                          &                          & 66.5          & 40.2          &                          &                          & 47.7          & 84.1          &                          &                          & 89.9                 \\
BoT\cite{luo2019strong}                                      & 53.7          & 62.2          & 48.2          &                          &                          & 56.7          & 51.4          &                          &                          & 61.3          & 40.5          &                          &                          & 49.6          & 74.7          &                          &                          & 81.3                 \\
DDAN\cite{chen2021dual}                                     & 59.0          & 63.1          & 52.3          & \multicolumn{1}{c}{60.6} & \multicolumn{1}{c}{71.8} & 56.4          & 54.5          & \multicolumn{1}{c}{62.7} & \multicolumn{1}{c}{74.9} & 58.9          & 50.6          & \multicolumn{1}{c}{62.1} & \multicolumn{1}{c}{73.8} & 55.7          & 78.5          & \multicolumn{1}{c}{85.3} & \multicolumn{1}{c}{92.5} & 81.5                 \\
DMG-Net\cite{bai2021person30k}                                  & 61.2          & 67.3          & 53.9          &                          &                          & 60.4          & 60.6          &                          &                          & 68.4          & 51.0          &                          &                          & 56.6          & 79.3          &                          &                          & 83.9                 \\
RaMoE\cite{dai2021generalizable}                                   & 61.5          & 69.1          & 56.6          &                          &                          & 64.6          & 57.7          &                          &                          & 67.3          & 46.8          &                          &                          & 54.2          & \textbf{85.0} &                          &                          & \textbf{90.2}        \\
MetaBIN\cite{choi2021meta}                                  & 66.0          & 73.6          & 59.9          & \multicolumn{1}{c}{78.4} & \multicolumn{1}{c}{82.8} & 68.6          & \textbf{74.2} & \multicolumn{1}{c}{89.7} & \multicolumn{1}{c}{92.2} & \textbf{81.0} & 48.4          & \multicolumn{1}{c}{70.3} & \multicolumn{1}{c}{77.2} & 57.9          & 81.3          & \multicolumn{1}{c}{95.0} & \multicolumn{1}{c}{97.0} & 87.0                 \\
BCaR\cite{tamura2020bcar}                                     & 67.5          &               & 65.8          &                          &                          &               & 70.2          &                          &                          &               & 52.8          &                          &                          &               & 81.3          &                          &                          & \multicolumn{1}{l}{} \\
GDNorm\cite{liu2022debiased}                                   & {\ul 68.9}    & {\ul 76.3}    & \textbf{66.1} & \multicolumn{1}{c}{83.5} &                          & \textbf{74.1} & {\ul 72.6}    & \multicolumn{1}{c}{89.3} &                          & 79.9          & {\ul 55.4}    & \multicolumn{1}{c}{73.8} &                          & {\ul 63.8}    & 81.3          & \multicolumn{1}{c}{94.0} &                          & 87.2                 \\
AMS                                          & \textbf{69.2} & \textbf{76.7} & {\ul 63.1}    & \multicolumn{1}{c}{81.7} & \multicolumn{1}{c}{86.6} & {\ul 71.7}    & 72.3          & \multicolumn{1}{c}{88.0} & \multicolumn{1}{c}{92.3} & {\ul 80.0}    & \textbf{58.2} & \multicolumn{1}{c}{77.3} & \multicolumn{1}{c}{82.0} & \textbf{66.8} & {\ul 83.0}    & \multicolumn{1}{c}{94.2} & \multicolumn{1}{c}{97.2} & {\ul 88.1}           \\ \hline
\end{tabular}
  \label{tab:sota}
\end{table*}

\textbf{Comparison under Protocol-2 and Protocol-3}. As shown in Table 3, our method is compared with DualNorm[37]\cite{jia2019frustratingly}, QAConv$_50$\cite{liao2020interpretable}, and M$^3$L\cite{zhao2021learning} under Protocol-2, and RaMoE under Protocol-3, respectively\cite{dai2021generalizable}, and GDNorm\cite{liu2022debiased} for comparison. On average, our method is 3.2\% and 6.0\% higher in Protocol-2 than the SOTA's mAP and R1, and 2.4\% and 2.9\% higher in Protocol-3, far superior to these methods. Specifically, On Market-1501, our method outperforms the SOTA methods by 4.1\% mAP and 4.0\% R1 under Protocol-2, and 3.0\% and 3.2\% under Protocol-3. On DukeMTMC-reID, our method improves R1 accuracy and mAP by 1.3\% and 2.6\% respectively under Protocol-2 and 1.7\% under Protocol-3. When tested on CUHK03, R1 accuracy and mAP are 1.7\% and 1.8\% higher than the SOTA methods under Protocol-2, and 2.3\% and 2.5\% higher under Protocol-3, respectively. Our method also improves R1 accuracy by 2.2\%/2.8\% and mAP by 7.5\%/4.4\% on MSMT17, respectively. The performance on these four large-scale Re-ID datasets demonstrates the strong domain generalization ability of AMS.

\begin{table*}[ht]
\tabcolsep=4.5pt
\small
\centering
\caption{Comparison of performance with the SOTA methods under Protocol-2 and Protocol-3(\%)}

\begin{tabular}{l|c|cc|cc|cc|cc|cc}
\hline
\multicolumn{1}{c|}{\multirow{2}{*}{Method}} & \multirow{2}{*}{Protocol}   & \multicolumn{2}{c|}{Market}   & \multicolumn{2}{c|}{Duke}     & \multicolumn{2}{c|}{CUHK03}   & \multicolumn{2}{c|}{MSMT17}   & \multicolumn{2}{c}{Avg}       \\ \cline{3-12}        \multicolumn{1}{c|}{}                        &                             & mAP           & R1            & mAP           & R1            & mAP           & R1            & mAP           & R1            & mAP           & R1            \\ \hline
DualNorm\cite{jia2019frustratingly}                                 & \multirow{5}{*}{Protocol-2} & {\ul 52.3}    & {\ul 78.9}    & {\ul 51.7}    & 68.5          & 27.6          & 28.8          & {\ul 15.4}    & {\ul 37.9}    & 36.8          & 53.5          \\
QAConv$_{50}$\cite{liao2020interpretable}                                 &                             & 35.6          & 65.7          & 47.1          & 66.1          & 21.0          & 23.5          & 7.5           & 24.3          & 27.8          & 44.9          \\
M$^3$L(ResNet-50)\cite{zhao2021learning}                           &                             & 48.1          & 74.5          & 50.5          & {\ul 69.4}    & 29.9          & 30.7          & 12.9          & 33.0          & 35.4          & 51.9          \\
M$^3$L(IBN-Net50)\cite{zhao2021learning}                           &                             & 50.2          & 75.9          & 51.1          & 69.2          & {\ul 32.1}    & {\ul 33.1}    & 14.7          & 36.9          & {\ul 37.0}    & {\ul 53.8}    \\
AMS                                          &                             & \textbf{56.4} & \textbf{82.9} & \textbf{53.0} & \textbf{72.0} & \textbf{33.8} & \textbf{34.9} & \textbf{17.6} & \textbf{45.4} & \textbf{40.2} & \textbf{58.8} \\ \hline
RaMoE\cite{dai2021generalizable}                                    & \multirow{3}{*}{Protocol-3} & 56.5          & 82.0          & 56.9          & 73.6          & 35.5          & 36.6          & 13.5          & 34.1          & 40.6          & 56.6          \\
GDNorm\cite{liu2022debiased}                                   &                             & {\ul 68.2}    & {\ul 86.5}    & {\ul 63.8}    & {\ul 78.2}    & {\ul 47.9}    & {\ul 48.6}    & {\ul 20.4}    & {\ul 48.1}    & {\ul 50.1}    & {\ul 65.4}    \\
AMS                                          &                             & \textbf{71.2} & \textbf{89.7} & \textbf{65.5} & \textbf{79.9} & \textbf{50.2} & \textbf{51.1} & \textbf{23.2} & \textbf{52.5} & \textbf{52.5} & \textbf{68.3} \\ \hline
\end{tabular}
\end{table*}

\textbf{Comparative analysis experiments with different modules}.
For the different combinations of instance normalization and group whitening mentioned in Fig.2, we compare their performance on large-scale datasets in detail in Table 4, where IN\&GW denotes Fig. 2(c), IN$|$XGW denotes Fig. 2(d), GW$|$XIN represents Fig. 2(e), GW$|$IN represents Fig. 2(b), IN$|$GW represents Fig. 2(a).
From Table 4, we can see that combinations of instance normalization with group whitening operation show the most superior performance. Among them, the methods that only use instance normalization or group whitening have a great improvement over the baseline methods, which shows the advantages of invariant methods in extracting domain-invariant features. Compared with the tandem method, the parallel method has a great decrease, since instance normalization is a narrow application of the whitening operation, and the parallel method emphasizes the normalization part and weakens the whitening part. In our experiments, the tandem method is further divided into applying group whitening first or instance normalization first. For the former, IN may fail due to the strong effect of whitening, which also reflects that the former has lower accuracy than the corresponding latter. As for the simultaneous output and input of the first operation to the second operation for feature extraction, analogous to the parallel method, this highlights role of the normalization operation in feature extraction, thereby reducing the influence of the whitening operation.

For the attention module, we also conduct comprehensive experiments on different combinations of channel attention and spatial attention for IN and GW, i.e. CA, SA, CASA for IN and CA, SA, CASA for GW at the same time, and integrate IN and GW with different attention. From Table 5, we can see that the experimental performance of applying CA to GW is stronger than the corresponding method of applying SA to GW, and the experimental performance of applying SA to IN is stronger. Furthermore, we can see that applying CASA to IN and GW is weaker than applying single attention, so we can learn that spatial attention is suitable for instance normalization operations, while channel attention is suitable for group whitening operations.

\begin{table*}[ht]
\tabcolsep=4.5pt
\small
\centering
\caption{Comparison experiment of different modules under Protocol-3(\%)}
\begin{tabular}{c|cc|cc|cc|cc|cc}
\hline
\multirow{2}{*}{Method} & \multicolumn{2}{c|}{Market}     & \multicolumn{2}{c|}{Duke}       & \multicolumn{2}{c|}{CUHK03}     & \multicolumn{2}{c|}{MSMT17}     & \multicolumn{2}{c}{Avg}         \\ \cline{2-11} 
                        & mAP            & R1             & mAP            & R1             & mAP            & R1             & mAP            & R1             & mAP            & R1             \\ \hline
baseline                & 64.85          & 85.57          & 60.69          & 76.03          & 40.69          & 42.14          & 16.68          & 41.84          & 45.73          & 61.40          \\
IN                      & 70.27          & \textbf{89.37} & 64.56          & {\ul 79.76}    & 48.45          & 49.43          & 22.27          & 51.20          & 51.39          & 67.44          \\
GW                      & {\ul 70.64}    & 88.87          & 64.22          & 79.44          & 48.53          & 50.07          & 20.65          & 48.19          & 51.01          & 66.64          \\
IN\&GW                   & 69.11          & 88.78          & {\ul 64.72}    & 79.40          & 47.81          & 49.43          & 22.02          & 51.09          & 50.92          & 67.18          \\
IN$|$XGW                  & 70.56          & {\ul 89.04}    & 64.50          & 79.31          & 48.12          & 49.57          & 22.28          & 51.39          & 51.37          & 67.33          \\
GW$|$XIN                  & 69.56          & 88.63          & 64.10          & 78.55          & 48.91          & 49.93          & 21.89          & 51.01          & 51.12          & 67.03          \\
GW$|$IN                   & 70.08          & 88.57          & 64.33          & 79.31          & {\ul 49.03}    & {\ul 50.29}    & {\ul 22.37}    & {\ul 51.58}    & {\ul 51.45}    & {\ul 67.44}    \\
IN$|$GW                   & \textbf{70.95} & 88.95          & \textbf{64.90} & \textbf{79.85} & \textbf{49.45} & \textbf{51.14} & \textbf{22.74} & \textbf{52.23} & \textbf{52.01} & \textbf{68.04} \\ \hline
\end{tabular}
\end{table*}

\begin{table*}[ht]
\tabcolsep=6pt
\small
\centering
 \caption{Comparison experiment of different attention modules under Protocol-3(\%)}
\begin{tabular}{l|cc|cc|cc|cc|cc}
\hline
\multicolumn{1}{c|}{\multirow{2}{*}{Method}} & \multicolumn{2}{c|}{Market}     & \multicolumn{2}{c|}{Duke}       & \multicolumn{2}{c|}{CUHK03}     & \multicolumn{2}{c|}{MSMT17}    & \multicolumn{2}{c}{Avg}        \\ \cline{2-11} 
\multicolumn{1}{c|}{}                        & mAP            & R1             & mAP            & R1             & mAP            & R1             & mAP           & R1             & mAP           & R1             \\ \hline
tandem IN CA, GW SA                          & 70.29          & 89.37          & 65.26          & 79.53          & 47.48          & 49.14          & 22.27         & 51.66          & 51.33         & 67.43          \\
tandem IN SA, GW CA                          & \textbf{71.17} & 89.67          & \textbf{65.45} & 79.89          & \textbf{50.18} & \textbf{51.14} & \textbf{23.20} & \textbf{52.47} & \textbf{52.50} & \textbf{68.29} \\
tandem IN CA, GW CA                          & 70.15          & 89.19          & 64.45          & 79.44          & 49.57          & 49.86          & 22.02         & 50.50           & 51.55         & 67.25          \\
tandem IN SA, GW SA                          & 70.91          & 89.34          & 64.91          & 79.67          & 49.29          & 50.86          & 22.32         & 51.32          & 51.86         & 67.80           \\
tandem IN SA, GW CASA                        & 70.38          & 89.40           & 65.15          & \textbf{80.48} & 48.55          & 50.93          & 22.50          & 51.27          & 51.65         & 68.02          \\
tandem IN CASA, GW CA                        & 69.80           & \textbf{89.73} & 65.42          & 80.07          & 49.02          & 49.86          & 22.58         & 52.02          & 51.71         & 67.92          \\
tandem IN CASA, GW CASA                      & 69.00          & 88.75          & 65.22          & 79.85          & 48.28          & 49.50          & 22.47         & 51.53          & 51.24         & 67.41          \\ \hline
\end{tabular}
\end{table*}

\subsection{Ablation Study}
We conduct a comprehensive ablation study to demonstrate the effectiveness of our approach and detailed components by generalizing performance on person Re-ID benchmarks in large-scale domains.

\textbf{Effectiveness of model components:} To investigate the effectiveness of each component in this method, we conduct ablation experiments, shown in Table 5. MS represents the multi-operation of instance normalization followed by group whitening, and A represents the attention method. AMS outperforms the baseline mAP by 8.76\%, which demonstrates that combination of the two invariant methods can effectively enhance the generalization ability of the model. On this basis, the mAP of the model established by the attention method is increased by 0.73\%, which verifies the effectiveness of the attention method.

\begin{table}[t]
\tabcolsep=6pt
\small
\centering
\caption{Ablation Study under Protocol-3(\%)}
\begin{tabular}{cc|cccc}
\hline
\multirow{2}{*}{A} & \multirow{2}{*}{MS} & \multicolumn{4}{c}{M+D+MT$\rightarrow$C3} \\ \cline{3-6} 
                   &                  & mAP    & mAP   & mAP   & mAP   \\ \hline
$\times$           & $\times$         & 40.69  & 42.14 & 62.86 & 72.86 \\
$\times$           & \checkmark       & 49.45  & 51.14 & \textbf{72.21} & \textbf{81.57} \\
\checkmark         & \checkmark       & \textbf{50.18}  & \textbf{51.14} & 71.50 & 80.71 \\ \hline
\end{tabular}
\end{table}
\textbf{Effectiveness of hyperparameter \textsl{g}:} To make the number of groups g be divisible by the number of channels (256, 512, 1024) of the three bottleneck layers, the number of groups can take 2, 4, 8, 16, 32, 64,  and 128. During the experiments, we found that when the number of groups is 2, 4, and 8, the loss function becomes NaN. Therefore, we analyze the grouping numbers 16, 32, 64 and 128 on the MS, i.e. the multi-operation of instance normalization followed by group whitening. In Table 6, we observe that in terms of mAP, when \textsl{g} increases from 16 to 128, it decreases by 0.91\%; when \textsl{g }increases from 16 to 64, R1 accuracy increases by 1.67\%. When \textsl{g} increases from 64 to 128, R1 accuracy drops by 1.13\%. However, the loss function with the number of groups 16 and 32 is NaN with applying the attention method, as for 64 and 128, 64 has 0.8\% higher mAP than 128 and 0.61\% higher R1, therefore we set \textsl{g} to be 64, and AMS achieves the best performance.

\begin{table}[t]
\tabcolsep=6pt
\small
\centering
\caption{Ablation experiment of group number g under Protocol-2 (\%)}
\begin{tabular}{c|c|cccc}
\hline
\multirow{2}{*}{$g$} & \multirow{2}{*}{Method} & \multicolumn{4}{c}{C3+D+MT$\rightarrow$M} \\ \cline{3-6} 
                   &                         & mAP    & R1    & R5    & R10   \\ \hline
16                 & \multirow{4}{*}{MS}     & \textbf{56.14}  & 81.50  & 91.98 & 94.86 \\
32                 &                         & 55.97  & 81.77 & 92.16 & \textbf{95.16} \\
64                 &                         & 55.76  & \textbf{83.17} & \textbf{92.64} & 95.10  \\
128                &                         & 55.23  & 82.04 & 91.78 & 94.42 \\ \hline
64                 & \multirow{2}{*}{AMS}    & \textbf{56.41}  & \textbf{82.90}  & 92.22 & \textbf{94.83} \\
128                &                         & 55.61  & 81.29 & \textbf{92.37} & \textbf{94.83} \\ \hline
\end{tabular}
\label{tab:hyper}
\end{table}

\section{Conclusion}
In this paper, we propose a novel attention-aware multi-operation module (AMS) for domain-generalized person re-identification, which can extract domain-invariant feature representations. Compared with the SOTA methods, the proposed method employs a tandem combination of instance normalization and group whitening to extract domain-invariant feature representations. Furthermore, we use an attention-aware approach to further enhance the generalization ability of the model. Extensive experiments demonstrate the effectiveness of our method for generalizing person Re-ID models in the unseen domain. Our method achieves the SOTA results on multiple benchmarks.

	\bibliographystyle{IEEEtran}
	\bibliography{references}
\end{document}